\documentclass[letterpaper, 10 pt, conference]{ieeeconf} 

\IEEEoverridecommandlockouts                              
\usepackage{times}
\usepackage{epsfig}
\usepackage{graphicx}
\usepackage{amsmath}
\usepackage{amssymb}
\usepackage{xargs} 

\usepackage{caption}
\usepackage{subcaption}
\usepackage[nolist]{acronym}
\usepackage{multirow}

\usepackage[pagebackref=true,breaklinks=true,colorlinks,bookmarks=false]{hyperref}

\begin{document}

\begin{acronym}
\acro{vo}[VO]{Visual Odometry}
\acro{cnn}[CNN]{Convolutional Neural Network}
\acro{rnn}[RNN]{Recurrent Neural Network}
\acro{mdn}[MDN]{Mixture Density Network}
\acro{lstm}[LSTM]{Long Short-Term Memory}
\end{acronym}

\title{\LARGE \bf MDN-VO: Estimating Visual Odometry with Confidence}

\author{Nimet Kaygusuz, Oscar Mendez, Richard Bowden \thanks{All authors are with the University of Surrey \tt\small \{n.kaygusuz, o.mendez, r.bowden\}@surrey.ac.uk}
}

\maketitle

\begin{abstract}
\ac{vo} is used in many applications including robotics and autonomous systems. However, traditional approaches based on feature matching are computationally expensive and do not directly address failure cases, instead relying on heuristic methods to detect failure. In this work, we propose a deep learning-based \acs{vo} model to efficiently estimate 6-DoF poses, as well as a confidence model for these estimates.
We utilise a \acs{cnn} - \acs{rnn} hybrid model to learn feature representations from image sequences. We then employ a \acf{mdn} which estimates camera motion as a mixture of Gaussians, based on the extracted spatio-temporal representations. 

Our model uses pose labels as a source of supervision, but derives uncertainties in an unsupervised manner. 
We evaluate the proposed model on the KITTI and nuScenes datasets and report extensive quantitative and qualitative results to analyse the performance of both pose and uncertainty estimation. Our experiments show that the proposed model exceeds state-of-the-art performance in addition to detecting failure cases using the predicted pose uncertainty.

\end{abstract}

\section{Introduction}
\label{section:introduction}

Traditional \ac{vo} approaches have been studied for decades and supported real world applications in robotics, computer vision and autonomous driving. Even though these algorithms perform well in ideal conditions, they are not robust, perform poorly in low texture environments and are prone to failure under fast motion \cite{scaramuzza2011visual}. They also do not have built-in reliability estimation which limits their ability to recover from failure cases. As traditional \ac{vo} algorithms do not provide a reliability measure, systems that depend upon their output, such as path planning \cite{zhu2017target}, vehicle state estimation \cite{weiss2013monocular} etc., are therefore susceptible to failure which can lead catastrophic outcomes.

Recent years have seen a move to pose estimation using learning based approaches \cite{wang2017deepvo, li2018undeepvo, yang2020d3vo}. Unlike traditional VO algorithms, learning-based methods exploit the availability of large scale datasets to learn from the data itself \cite{krizhevsky2012imagenet}. This enables them to be robust to conditions such as low texture areas and challenging lighting conditions without requiring accurate camera calibration \cite{chen2020survey}. 

\begin{figure}[t]
\begin{center}
   \includegraphics[width=0.8\linewidth]{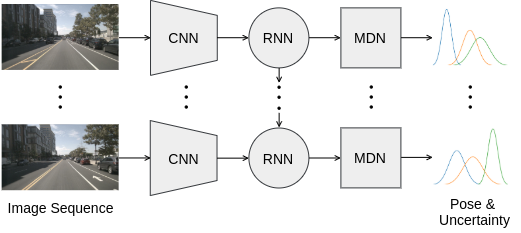}
\end{center}
   \caption{An overview of the proposed monocular VO and uncertainty approach using MDN.}
\label{fig:onecol}
\end{figure}

Motivated by the recent success of deep learning based approaches and to address the pose estimation reliability, in this work we present a novel deep learning based VO estimation approach that includes uncertainty. We work on raw images and extract image representations using a \acf{cnn} \cite{krizhevsky2012imagenet}. We then model the vehicle motion in the image sequences using an \acf{rnn} \cite{hochreiter1997long}. Unlike most approaches, which regress a single pose \cite{mohanty2016deepvo, wang2017deepvo}, we employ a \acf{mdn} \cite{bishop1994mixture} to regress a mixture model of 6-DoF poses (see Figure~\ref{fig:onecol}). This approach enables us to estimate pose and uncertainty using the predicted mixture model characteristics which are means $(\mu)$, standard deviations $(\sigma)$ and mixture coefficients $(\alpha)$. 

We evaluate our approach on two challenging public datasets, namely KITTI \cite{geiger2013vision} and nuScenes \cite{caesar2020nuscenes}. We report quantitative and qualitative pose estimation and VO reliability results. We show that uncertainty estimates negatively correlate with pose estimation accuracy. This indicates that the proposed approach can successfully predict \ac{vo} reliability without requiring additional supervision.

The contributions of this paper can be summarised as:
\begin{itemize}
\item We propose a learning-based \ac{vo} estimation approach that employs a mixture of probability distributions to enable uncertainty estimation.
\item We demonstrate state-of-the-art performance on the KITTI and the recently released nuScenes datasets, which includes challenging scenarios such as night-time and fast motion.
\item We provide extensive analysis for uncertainty estimations and demonstrate how they reflect pose estimation failures.
\end{itemize}

The rest of the paper is structured as follows: In Section \ref{section:literature-review} we discuss traditional and learning based approaches to \ac{vo}. We then introduce our approach in Section \ref{section:methodology}. In Section \ref{section:experiments} we share our experimental setup and report qualitative and quantitative results. We conclude in Section \ref{section:conclusion} by discussing our findings and possible future work. 

\section{Related Work}
\label{section:literature-review}

Traditional feature based methods to VO estimation have been studied for the last two decades \cite{nister2004visual, mur2015orb}. They estimate the motion of the camera by extracting and matching a group of hand crafted feature points (e.g. SIFT \cite{lowe2004distinctive}, SURF \cite{bay2006surf} or ORB \cite{rublee2011orb}) on consecutive frames. However, these approaches are prone to failure in scenarios where there are insufficient features, such as in low texture environments or poor illumination \cite{scaramuzza2011visual}. To address this issue, Newcombe et al. \cite{newcombe2011dtam} proposed a dense direct approach, which estimates the motion by optimising image pixel intensities. However, this approach requires extensive computational power. To mitigate the heavy computational requirements, sparse direct methods have been introduced. Engel et al. \cite{engel2017direct} proposed only using image pixels which have high gradient intensities. Compared to feature based approaches, direct methods are more suited to low texture environments but require a good initialisation and are sensitive to rapid motion and photometric change.

More recently, deep learning based approaches have dominated computer vision \cite{krizhevsky2012imagenet}. 
Inspired by this, the field has adapted \ac{cnn}s to extract and match feature points and to estimate \ac{vo} \cite{wang2020approaches}.
DeTone et al. \cite{detone2018superpoint} present a learning-based approach to detect interest point detectors and descriptors.
Sarlin et al. \cite{sarlin2020superglue} propose a neural network to learn to match two sets of pre-existing features. End-to-end \ac{vo} approaches have also been studied:
Mohanty et al. \cite{mohanty2016deepvo} propose combining two \ac{cnn}s to extract features from sequential images. They estimate pose by concatenating features and passing them through fully connected layers. To enhance the temporal modelling capabilities, Wang et al. \cite{wang2017deepvo, wang2018end} propose using \ac{rnn}s to model changes in image features and achieve competitive results to traditional VO based approaches. 

Most deep learning based VO approaches are trained in a supervised manner, which requires labelled data. However, collecting and annotating large amounts of data is a laborious task. Alternatively, unsupervised learning based approaches \cite{li2018undeepvo, yin2018geonet, zhan2018unsupervised, zhao2018learning} have the ability to exploit vast amounts of unlabelled data. Zhou et al. \cite{zhou2017unsupervised} use view synthesis and learning depth and pose estimation together. They use a photometric error based loss function to train their networks. However, their approach is not able to recover the global scale. To solve this issue, Zhan et al. \cite{zhan2018unsupervised} and Li et al. \cite{li2018undeepvo} propose using stereo image pairs to estimate scaled VO in an unsupervised manner. Li et al. \cite{li2020self} present a meta-learning algorithm for a better adaptation to unseen environments.
Although unsupervised VO methods achieve promising results and allow the models to be trained on a large variety of unlabelled data, their current performance is lower than supervised methods. 

More recently Yang et al. \cite{yang2020d3vo} proposed a hybrid approach, which combines unsupervised deep learning and classical direct VO estimation methods. They estimate depth, pose and photometric uncertainty from images. They then feed these outputs to a direct VO estimation method \cite{engel2017direct} to obtain the final trajectories.

Uncertainty estimation has been studied for similar tasks, such as camera localisation. Kendall et al. \cite{kendall2016modelling} model camera localisation uncertainty by utilising dropout at test time. However this approach can be considered as model uncertainty estimation \cite{chen2020survey}, whereas in this work we approach uncertainty estimation as a sensor reliability metric, which can ultimately be used for sensor fusion.

Inspired by the recent success of deep learning based VO models, in this work we propose a novel architecture where we utilise \ac{mdn}s \cite{bishop1994mixture} to regress the 6-DoF poses as a mixture of distributions, which allows the network to learn multiple modes from the data.

\begin{figure*}[!ht]
\begin{center}
   \includegraphics[width=0.80\linewidth]{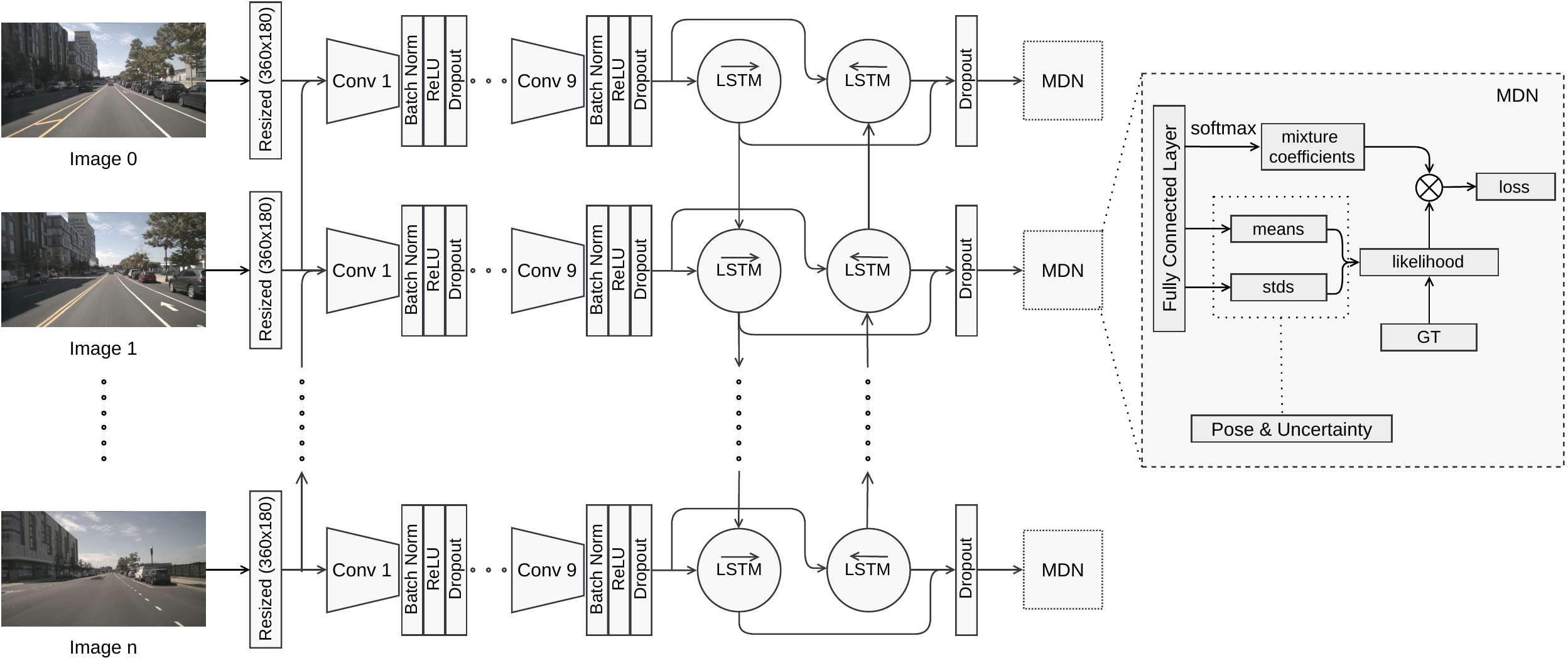}
\end{center}
   \caption{Architecture of the proposed VO and uncertainty estimation framework, consisting of visual feature extractor, temporal modelling and mixture model based pose and uncertainty regression.}
\label{fig:methodology}
\vspace{-0.1cm}
\end{figure*}

\section{Methodology}
\label{section:methodology}

In this section, we introduce the proposed end-to-end approach for estimating visual odometry and its confidence/uncertainty. Our model takes in a sequence of images, $\mathcal{V} = \{I_0, ..., I_T\}$, with $(T+1)$ number of frames, and predicts a mixture of distributions, $\mathcal{G}$, with $M$ components, that are most likely to produce the ground truth relative poses, $Y = \{y_{(0, 1)}, ..., y_{(T-1,T)}\}$. We use Gaussian distributions, $\mathcal{N}$, as mixture components to model relative poses. Thus, we can notate our mixture model $\mathcal{G}$ as:

\begin{equation}
\mathcal{G} = \{\alpha^1_t\mathcal{N}_t(\mu^1, \sigma^1), ..., \alpha^M_t \mathcal{N}_t(\mu^M, \sigma^M)\}^{t=1:T}    
\end{equation}

We break the proposed architecture into three main modules, namely \emph{Visual Feature Extraction}, \emph{Temporal Modelling} and \emph{Mixture Density Estimation}. In the visual feature extraction module, we extract features from consecutive image pairs using a \ac{cnn}. We then pass the extracted features to the temporal modelling layer, which utilises an \ac{rnn} to model the changes in visual features over time. Finally, the \ac{rnn} outputs are fed into the mixture density estimation module, which estimates the relative poses for each frame with respect to its predecessor. An overview of the proposed approach can be seen in Figure \ref{fig:methodology}. In the rest of this section we describe each component in detail.

\subsection{Visual Feature Extractor}
Traditional VO approaches start by extracting geometric features from image sequences. By matching the feature points between image pairs, these models estimate the camera motion. Following the same concept, learning based approaches \cite{mohanty2016deepvo, wang2017deepvo}, propose learning representations of distinctive image features using \ac{cnn}s. 

We develop our \ac{cnn} backbone structure similar to Flownet \cite{dosovitskiy2015flownet}. It is designed to learn the features that are most suitable to represent image motion in the context of optical flow, which is obviously a related task to VO estimation. The network consists of $9$ convolutional layers with a receptive filter size of $7x7$ for the first layer, $5x5$ for the next two and $3x3$ for the rest. We utilise batch normalisation \cite{ioffe2015batch}, Leaky ReLU \cite{maas2013rectifier} and Dropout \cite{srivastava2014dropout} after each convolutional layer.

Given an image sequence, $\mathcal{V}$, our model starts by concatenating consecutive image pairs $(I_{t-1}, I_{t})$ and extracting features, $f_t$ as:

\begin{equation}
    f_{t} = \mathrm{CNN(} [I_{t-1}, I_{t}] \mathrm{)}
\end{equation}

where $[.]$ is a concatenation operation over the image colour channels. Extracted image features $\{f_1, ..., f_{T}\}$ are then passed to the next module, which learns to model the temporal changes in the latent space. 

\subsection{Temporal Modelling}

We track the changes of distinguishable patterns on the image plane to estimate the camera motion \cite{scaramuzza2011visual}. Unlike commonly used hand-crafted features \cite{shi1994good, harris1988combined, lowe2004distinctive, bay2006surf, rosten2006machine}, in this work we utilise learnt image representations, and model their changes over time using \ac{rnn}s. 

Given image features, $\{f_1 ..., f_{T}\}$, \ac{rnn} produces outputs $r_t$, for each frame $I_t$ as:

\begin{equation}
    r_t, h_t = \mathrm{RNN(}f_{t}, h_{t-1} \mathrm{)}
\end{equation}

where $h_t$ is the hidden states of the \ac{rnn} units after producing $r_t$, and $h_0$ is a zero vector. 

We employ \ac{lstm} units \cite{hochreiter1997long} as our \ac{rnn} structure, which has been shown to be successful in modelling long-term dependencies. We utilise a bi-directional LSTM, to enhance the representation by using a small temporal window of past frames.

The LSTM outputs, $\{r_1, ..., r_{T}\}$, are then passed to the next module, which estimates the mixture of relative pose distributions, conditioned on these features.

\subsection{Mixture Density Estimation}

Unlike other deep learning based VO methods \cite{mohanty2016deepvo, wang2017deepvo}, which directly regresses a 6-DOF pose given an image sequence, our approach estimates a mixture model over the 6-DoF pose by using an \ac{mdn}.

The \ac{mdn} module takes temporally modelled features $\{r_1, ..., r_{T}\}$ and outputs the parameters of a mixture density model with $M$ components for each time step $t$, namely means $\{\mu_1, ..., \mu_M\}^t$, standard deviations $\{\sigma_1, ...,\sigma_M\}^t$ and mixture coefficients $\{\alpha_1, ...,\alpha_M\}^t$. We use a fully connected layer to predict these parameters from LSTM outputs.

Given a set of temporally modelled features, $\{r_1, ..., r_{T}\}$, we model the conditional probability of pose change, $y_{(t-1,t)}$, at time $t$ as:

\begin{equation}
    p\left(y_{(t-1,t)}|r_t\right) = \sum_{i=1}^{M}
    \alpha_{i} \left(r_t\right) \phi \left(y_{(t-1,t)}|r_t\right)
\end{equation}

where M is the number of mixture units, $\alpha_i\left(r_t\right)$ is the mixture coefficient which represent the probability of the pose, $y_{(t-1,t)}$, being in the $i^{th}$ component given $r_t$. The conditional density, $\phi \left(y_{(t-1,t)}|r_t\right)$, of the pose $y_{(t-1,t)}$, for the $i^{th}$ component, can be expressed as a Gaussian distribution:

\begin{equation}
    \phi \left(y_{(t-1,t)}|r_t\right) = \frac{1}
    {\sigma_{i} \left(r_t\right) \sqrt{2{\pi}}} 
    e^{\left\{-\frac{{\|y_{(t-1,t)} - \mu_i\left(r_t\right)\|}^2}
    {2\sigma_i\left(r_t\right)}\right\}}
\end{equation}

where $\mu_i\left(r_t\right)$ and $\sigma_i\left(r_t\right)$ show the mean and standard deviation of the $i^{th}$ mixture.

We can calculate the likelihood of a pose estimate, $y_t$, with $K$ variables, i.e. $K=6$ for 6 degree of freedom pose, given $r_t$, as the product of the likelihood of each variable as:

\begin{equation}
    \mathcal{L}_t = \prod_{k=1}^K p\left(y_{(t-1,t)}[k]|r_t\right)
\end{equation}

where $[k]$ is an indexing operation to access $k^{th}$ variable of the pose. 

The average likelihood for a sequence with $T$ time steps, $\mathcal{L}$, is then calculated by summation over the time axis $t$ as:

\begin{equation}
    \mathcal{L} = \frac{1}{T} \sum_{t=1}^T \mathcal{L}_t
\end{equation}

We train our network to maximise the likelihood. Hence, we use the negative log likelihood as our error signal: 
\begin{equation}
    E = \mathrm{-log}(\mathcal{L})
\end{equation}

We use ground truth relative poses, ${Y = \{y_{(0, 1)}, ..., y_{(T-1,T)}\}}$, to train our network. During inference, we derive the relative poses and uncertainties using the mean and standard deviations of the mixture components, respectively. We recover absolute pose, $\mathcal{Y}_T$, by accumulating the relative poses over time as:

\begin{equation}
    \mathcal{Y}_{T}=\textbf{y}_{(T-1,T)}...\textbf{y}_{(1,2)}\textbf{y}_{(0,1)}
\end{equation}

where $\textbf{y}_{(t-1,t)}$ is the $4x4$ homogeneous transformation matrix representation of pose change $y_{(t-1,t)}$.

\section{Experiments}
\label{section:experiments}

We evaluate our approach on the KITTI \cite{geiger2013vision} and nuScenes \cite{caesar2020nuscenes} datasets and report quantitative and qualitative experiment results.

\textbf{KITTI} is one of the most popular autonomous driving datasets. It provides rectified camera images and ground truth motion of the vehicle. The driving scenarios include fast motions and sharp turns, presenting a challenging benchmark for egomotion estimation.

\textbf{nuScenes} is a recently released large-scale autonomous driving dataset which contains significant environmental variation. Compared to the KITTI dataset, nuScenes contains more varied sequences, including night time driving and rainy weather. We believe these challenging driving scenarios and environmental conditions present a good baseline for evaluating the necessity of \ac{vo} uncertainty estimation. 

\subsection{Implementation Details}
Our network is implemented using the PyTorch framework and trained on an NVIDIA TITAN X GPU. The training of the network takes approximately $40$ epochs to converge. We used the Adam optimiser with parameters $(\beta_1=0.9, \beta_2=0.999)$. We utilise a plateau learning rate scheduler with a starting learning rate of $10^{-3}$, patience of $8$ and decay factor of $0.7$. We also use dropout with a rate of $0.1$ on \ac{cnn} layers and a $0.2$ drop rate on LSTM layers to prevent over-fitting. We use pre-trained Flownet weights \cite{dosovitskiy2015flownet} to initialise \ac{cnn} backbone. For the rest of the parameters we used Xavier initialisation.
We use the \textit{evo} python package \cite{grupp2017evo} to measure the performance of our approach, using relative pose error (RPE).
 
\subsection{Experiments on the KITTI Dataset}

In our first set of experiments, we compared the proposed approach against the state-of-the-art on the KITTI dataset. We only considered the left colour camera as input to our network. For training, we used sequences 00, 01, 02, 04, 08, 09 and tested our model using sequences 03, 05, 06, 07, 10.

\begin{table*}[!ht]
\centering
\caption{Quantitative Results on the KITTI Dataset}
\begin{tabular}{c|clc|clc|clc}
\hline
\multirow{2}{*}{Sequence}     & \multicolumn{3}{c|}{ORB-SLAM} & \multicolumn{3}{c|}{DeepVO} & \multicolumn{3}{c}{\textbf{MDN-VO (ours)}} \\ \cline{2-10} 
     & RMSE          & Max              & Mean $\pm$ std            & RMSE          & Max           & Mean $\pm$ std            & RMSE               & Max              & Mean $\pm$ std           \\ \hline
03   & \textbf{0.03} & \textbf{0.20}    & \textbf{0.03 $\pm$ 0.02}  & 0.08          & 0.23          & 0.08 $\pm$ 0.04           & 0.12               & 0.41             & 0.11 $\pm$ 0.06          \\
05   & 0.25          & 0.67             & 0.20 $\pm$ 0.15           & 0.24          & 0.57          & 0.21 $\pm$ 0.12           & \textbf{0.16}      & \textbf{0.36}    & \textbf{0.15 $\pm$ 0.08} \\
06   & 0.34          & 0.65             & 0.30 $\pm$ 0.17           & \textbf{0.16} & \textbf{0.31} & \textbf{0.14 $\pm$ 0.08}  & 0.20               & 0.45             & 0.18 $\pm$ 0.09          \\
07   & 0.17          & \textbf{0.35}    & 0.13 $\pm$ 0.09           & 0.14          & \textbf{0.35} & 0.12 $\pm$ 0.07           & \textbf{0.08}      & 0.51             & \textbf{0.07 $\pm$ 0.05} \\
10   & 0.30          & 0.95             & 0.23 $\pm$ 0.20           & 0.21          & 0.47          & 0.19 $\pm$ 0.08           & \textbf{0.14}      & \textbf{0.32}    & \textbf{0.13 $\pm$ 0.06} \\ \hline
mean & 0.22          &  0.56            & 0.18  $\pm$ 0.13          & 0.17          & \textbf{0.39} & 0.15 $\pm$ 0.08           & \textbf{0.14}      & 0.41             & \textbf{0.13 $\pm$  0.07}                     \\ \hline
\end{tabular}
\label{table:kitti}
\end{table*}

We compare our approach with monocular ORB-SLAM \cite{mur2015orb} and DeepVO \cite{wang2017deepvo}. We ran ORB-SLAM without loop closure, to make it more comparable to the proposed approach in the context of VO. 
Unlike the proposed approach, monocular ORB-SLAM estimates the trajectory up to a scale thus we aligned its trajectory with ground truth using \cite{umeyama1991least}. We used the PyTorch implementation of DeepVO\footnote{https://github.com/ChiWeiHsiao/DeepVO-pytorch/}. 

As can be seen in Table~\ref{table:kitti}, the proposed approach (MDN-VO) outperforms both DeepVO and ORB-SLAM approaches in overall mean RMSE score. ORB-SLAM mostly suffers on manoeuvres that include sharp turns, as the algorithm relies on maintaining sufficient 3D-to-2D feature matches to obtain scaled poses between consecutive frames.

Overall, DeepVO performs better than ORB-SLAM but the proposed approach (MDN-VO) yields the best results. We believe this is due to modelling the target pose changes with a mixture model. A network trained by least squares approximates the conditional averages of the target data. However, simply modelling the mean is insufficient for multimodal distributions \cite{bishop1994mixture}. Thus, using a mixture model enables our model to learn multiple modes from the training data, overcoming the limitations of direct regression.

\begin{figure}[!ht]
\begin{center}
\begin{subfigure}[b]{0.9\linewidth}
\centering
   \includegraphics[width=0.7\columnwidth]{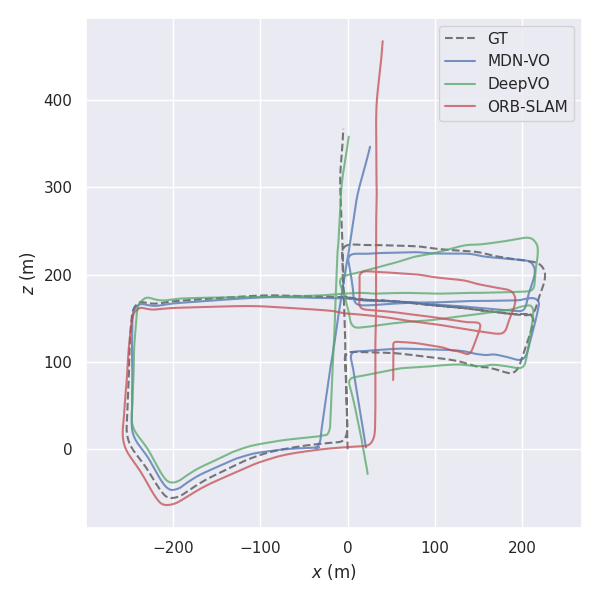}
   \caption{Sequence 05}
\end{subfigure}
\begin{subfigure}[b]{0.9\linewidth}
\centering
   \includegraphics[width=0.7\columnwidth]{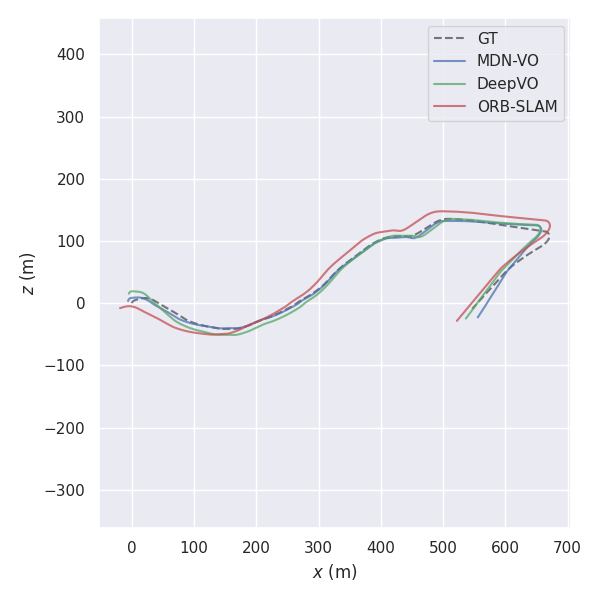}
    \caption{Sequence 10}
\end{subfigure}
\end{center}
   \caption{Estimated Trajectories on the KITTI Dataset.}
\label{fig:kitti:traj}
\vspace{-0.1cm}
\end{figure}

We also share aligned trajectories for the test sequences $5$ and $10$ in Figure~\ref{fig:kitti:traj}. As can be seen, ORB-SLAM performs drastically worse than learning based approaches in sequence $5$. We believe this is caused by the inconsistencies in estimated local scales due to the lack of global loop closure. As the proposed approach learns to estimate scale during training as a prior, we do not suffer from this issue. 

\begin{figure*}[!ht]
\begin{center}
\begin{subfigure}[b]{0.27\linewidth}
\centering
   \includegraphics[width=1.0\linewidth]{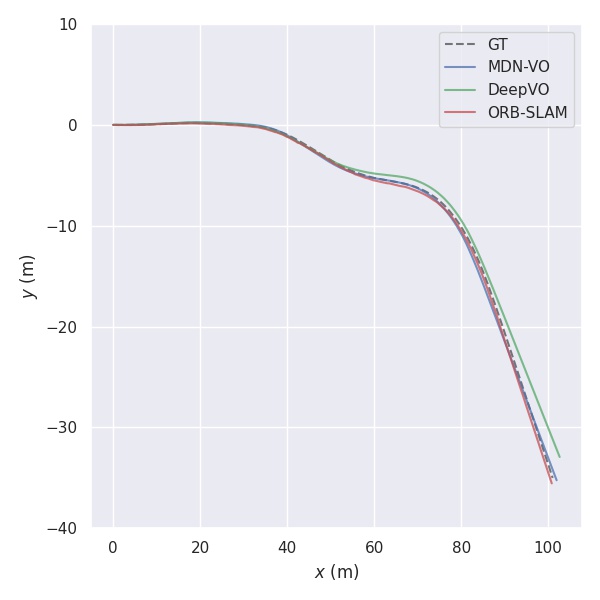}
   \hspace*{0.5cm}
   \includegraphics[width=0.75\linewidth]{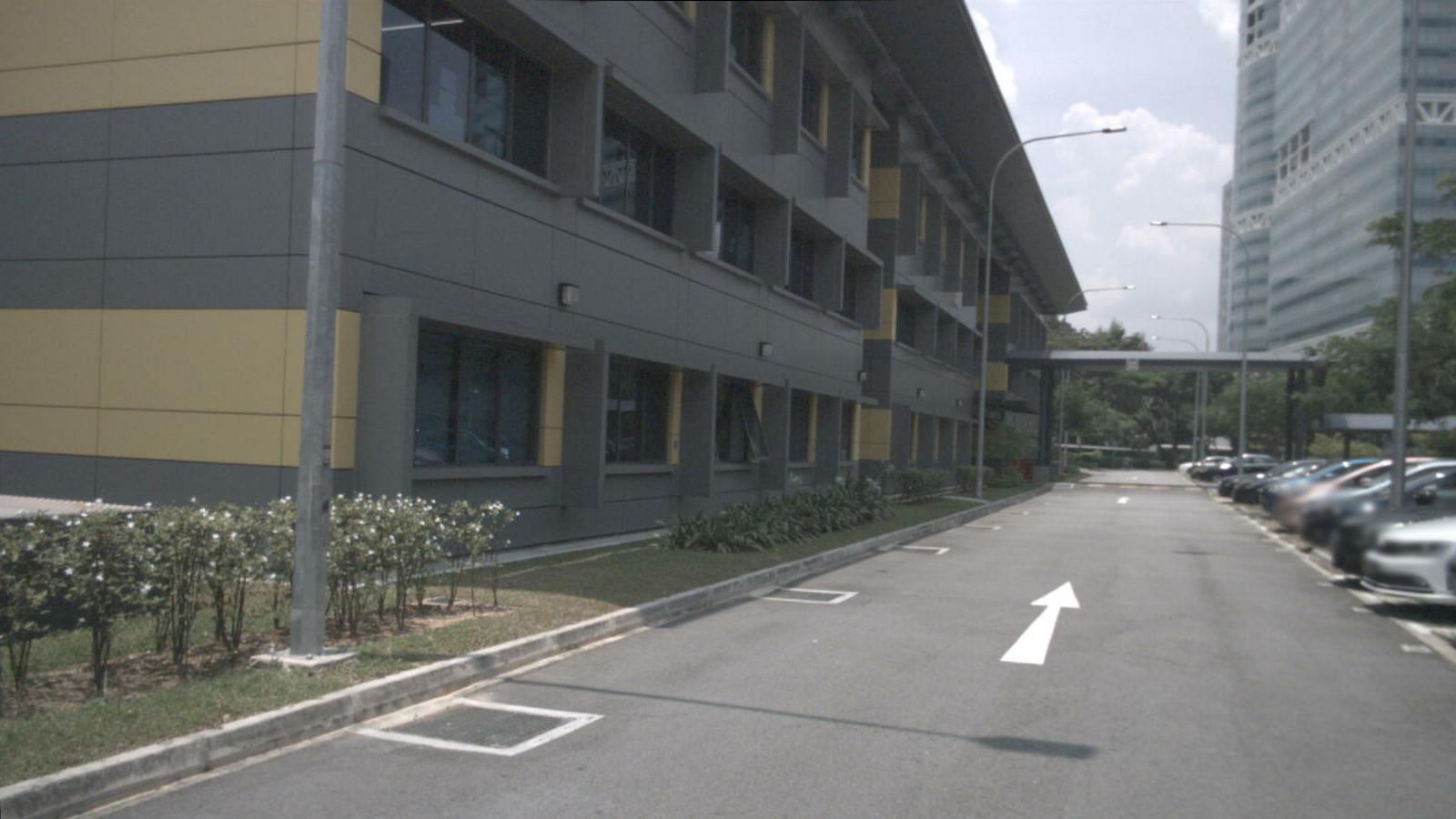}
   \caption{Daylight / scene-0972}
   \label{fig:nuscenesresult:a}
\end{subfigure}
\begin{subfigure}[b]{0.27\linewidth}
\centering
   \includegraphics[width=1.0\linewidth]{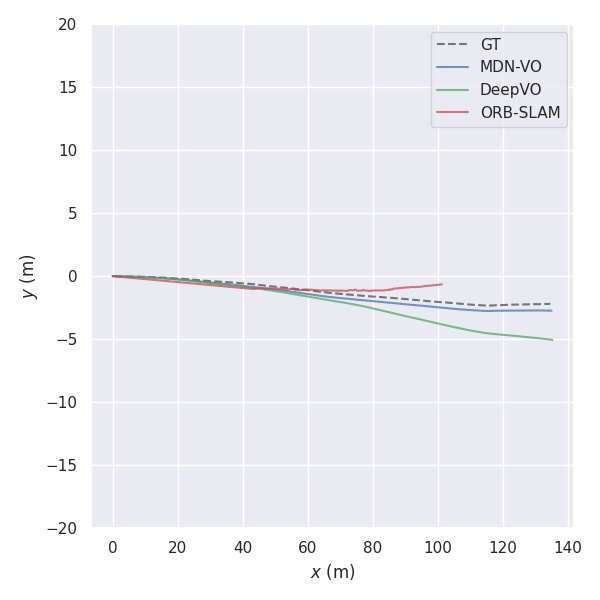}
   \hspace*{0.45cm}
   \includegraphics[width=0.75\linewidth]{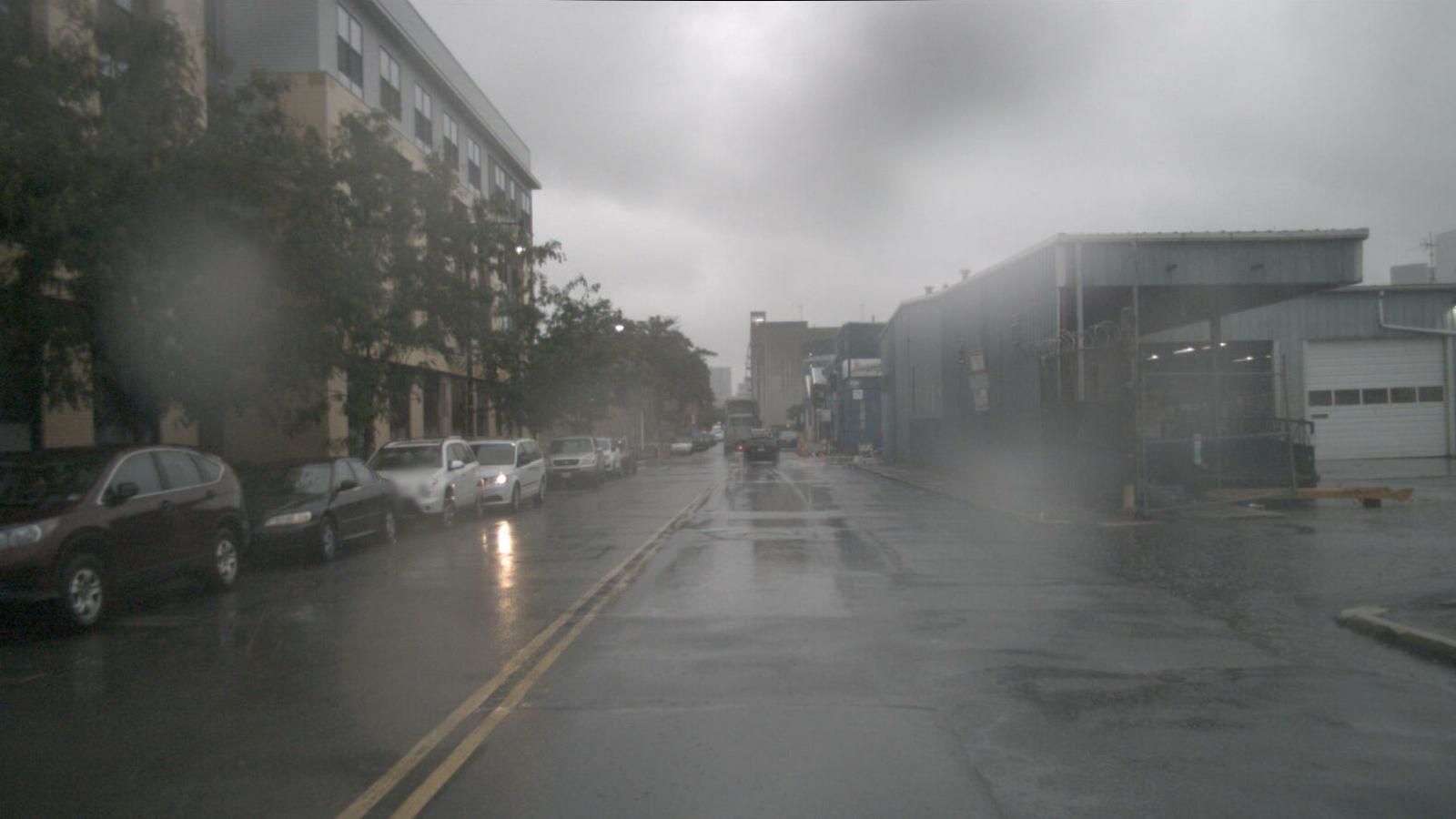}
   \caption{Rain / scene-0572}
   \label{fig:nuscenesresult:c}
\end{subfigure}
\begin{subfigure}[b]{0.27\linewidth}
\centering
   \includegraphics[width=1.0\linewidth]{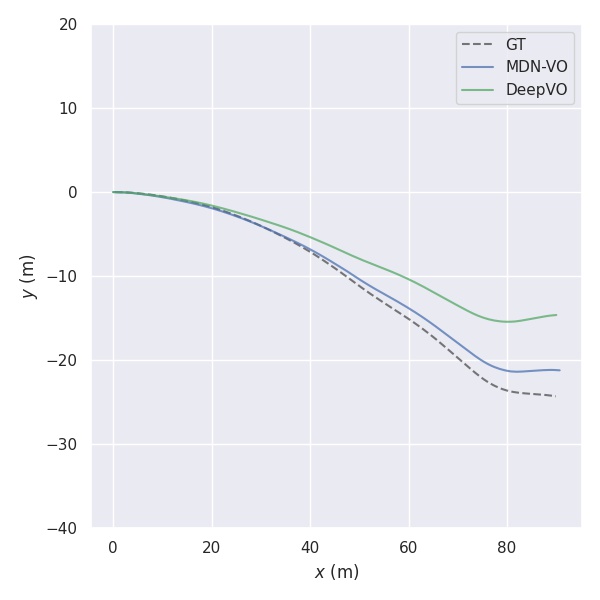}
   \hspace*{0.46cm}
   \includegraphics[width=0.75\linewidth]{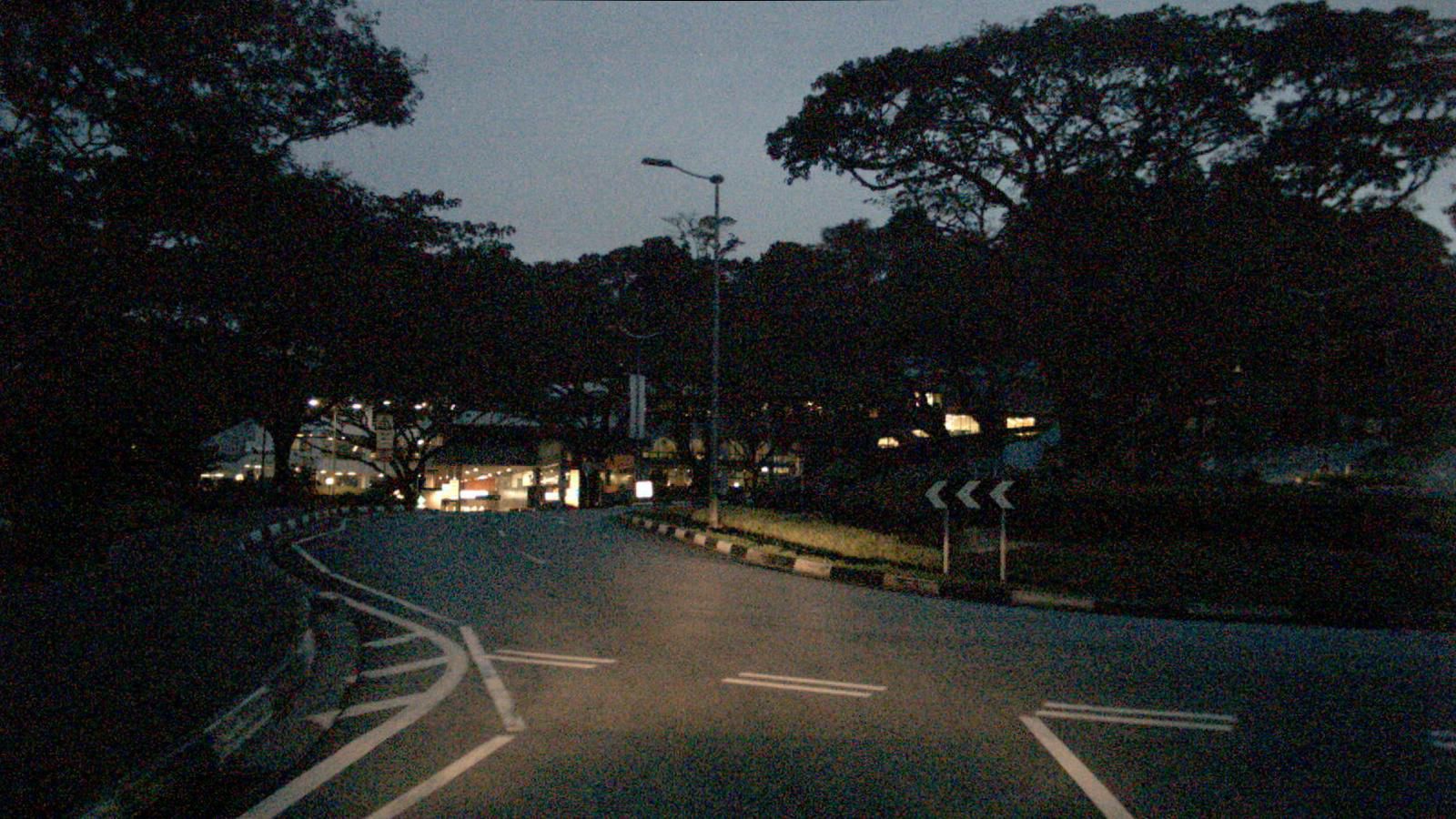}
    \caption{Night / scene-0999}
    \label{fig:nuscenesresult:d}
\end{subfigure}
\end{center}
   \caption{Estimated trajectories (top) and a sample frame (bottom) from the corresponding nuScenes sequences.}
\label{fig:nuscenesresult}
\end{figure*}

\begin{figure}[!ht]
\begin{center}
\begin{subfigure}[b]{0.49\linewidth}
\centering
   \includegraphics[width=1.0\linewidth]{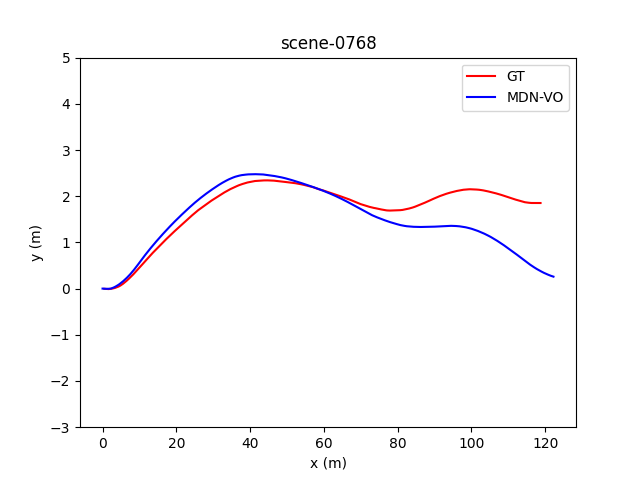}
   \caption{Trajectory}
   \label{fig:nuscenesuncertainty:a}
\end{subfigure}
\begin{subfigure}[b]{0.49\linewidth}
\centering
   \includegraphics[width=1.0\linewidth]{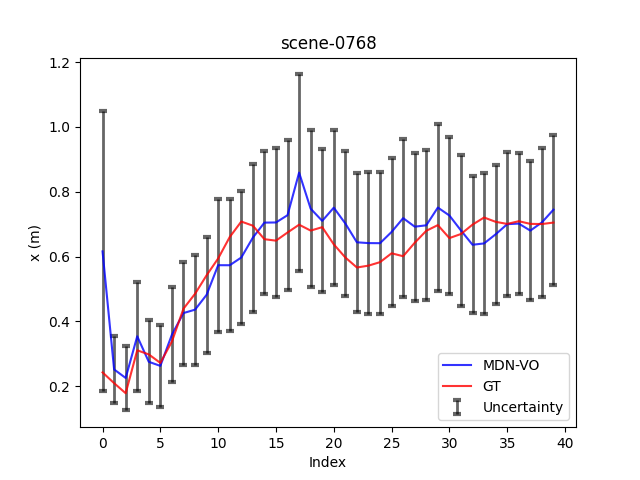}
    \caption{Uncertainty on x-position}
    \label{fig:nuscenesuncertainty:b}
\end{subfigure}
\begin{subfigure}[b]{0.49\linewidth}
\centering
   \includegraphics[width=1.0\linewidth]{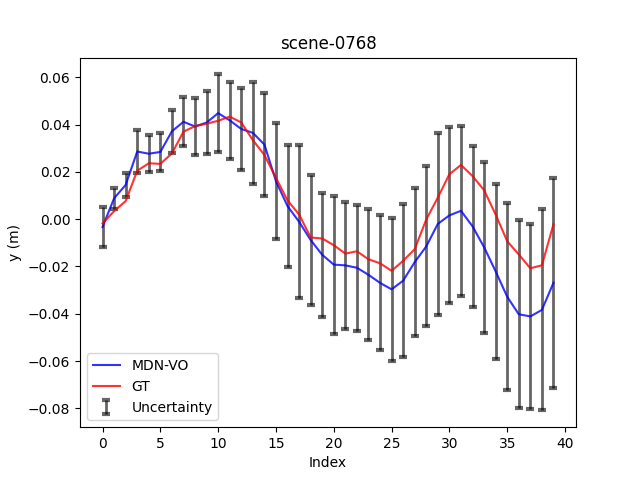}
   \caption{Uncertainty on y-position}
   \label{fig:nuscenesuncertainty:c}
\end{subfigure}
\begin{subfigure}[b]{0.49\linewidth}
\centering
   \includegraphics[width=1.0\linewidth]{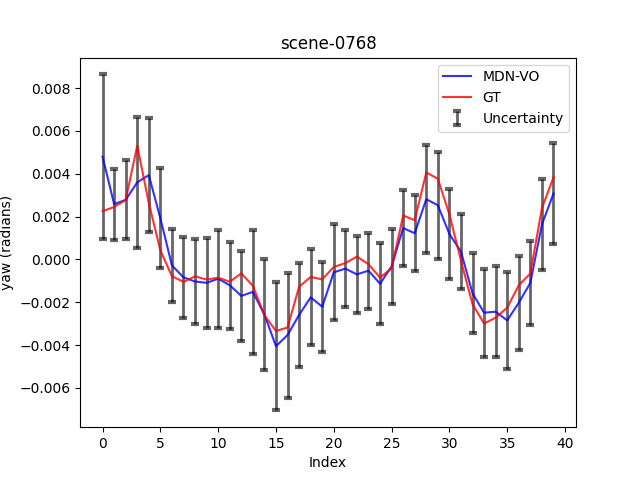}
    \caption{Uncertainty on yaw angle}
    \label{fig:nuscenesuncertainty:d}
\end{subfigure}
\end{center}
   \caption{Uncertainty estimation on a test sample from nuScenes dataset (scene-0768).}
\label{fig:nuscenesuncertainty}
\end{figure}

\begin{table*}[h]
\centering
\caption{Quantitative Results on the nuScenes Test Sequences}
\begin{tabular}{c|ccc|ccc|ccc}
\hline
\multirow{2}{*}{Condition} & \multicolumn{3}{c|}{ORB-SLAM} & \multicolumn{3}{c|}{DeepVO} & \multicolumn{3}{c}{MDN-VO (ours)} \\ \cline{2-10} 
          & RMSE & Max   & Mean $\pm$ std  & RMSE & Max  & Mean $\pm$ std  & RMSE          & Max  & Mean $\pm$ std           \\ \hline
Daylight  & 0.40 & 3.28  & 0.12 $\pm$ 0.38 & 0.09 & 0.54            & 0.06 $\pm$ 0.07 & \textbf{0.07} & \textbf{0.53} & \textbf{0.04 $\pm$ 0.05} \\
Rain      & 0.76 & 10.19 & 0.14 $\pm$ 0.74 & 0.07 & \textbf{0.39}   & 0.05 $\pm$ 0.05 & \textbf{0.06} & 0.43          & \textbf{0.04 $\pm$ 0.05} \\
Night     & -    & -     & -               & 0.12 & 0.74            & 0.09 $\pm$ 0.09 & \textbf{0.11} & \textbf{0.73} & \textbf{0.07 $\pm$ 0.08} \\ \hline
\end{tabular}
\label{table:nuscenes}
\end{table*}

\subsection{Experiments on the nuScenes Dataset}

Next, we used the nuScenes dataset to evaluate our approach and examine its effectiveness in difficult conditions. As the nuScenes dataset does not have a defined VO benchmark protocol, we define our own data split which includes scenarios from the challenging environmental conditions, such as direct sunlight, lack of illumination etc, in both training and validation sets.

In Figure~\ref{fig:nuscenesresult}, we share the estimated trajectories in three different weather/lighting conditions, namely daylight, rain and night time. The estimated trajectories and the ground truth can be seen in the top row, while we display sample frames from the corresponding sequences in the bottom row. As can be seen, the proposed method (MDN-VO) successfully produces trajectories for all sequences. However, ORB-SLAM failed to initialise in the night time scenarios, hence we could not report ORB-SLAM's qualitative (See Figure~\ref{fig:nuscenesresult:d}) and quantitative (See Table ~\ref{table:nuscenes}) results for these scenarios. Compared to ORB-SLAM and DeepVO, our method (MDN-VO) achieves more accurate trajectory estimates, even in challenging conditions such as rain and night scenes. Qualitative results are reflected in our quantitative experiments, where we use the Relative Pose Error (RPE) as our error metric and report performance of each method with respect to different weather conditions. As can be seen in Table~\ref{table:nuscenes}, the proposed approach surpasses the performance of both ORB-SLAM and DeepVO models.

We investigate uncertainty estimation of the proposed method in detail on a sample sequence, namely scene-0768. We plot the estimated trajectory against ground truth in Figure~\ref{fig:nuscenesuncertainty:a}. As can be seen, the error between the estimated trajectory and ground truth increases towards the end of the trajectory. To investigate if this result is captured by the uncertainty estimation, we visualise the relative pose estimations and the ground truth between consecutive frames along with the corresponding uncertainty estimations. We plot x, y and yaw angles separately in Figure~\ref{fig:nuscenesuncertainty:b}, Figure~\ref{fig:nuscenesuncertainty:c}, Figure~\ref{fig:nuscenesuncertainty:d}, respectively, to give the reader more insight.
As can be seen, the error between ground truth and our model's pose estimations are bounded by the uncertainty intervals ($3\sigma$). While the error in y estimate increases, the uncertainty estimate also increases. This shows that the failure on the $y$ estimate is captured by the model and is reflected to the $y$ uncertainty estimates (See Figure~\ref{fig:nuscenesuncertainty:c}). These results show that failure cases can be captured by the proposed model's uncertainty estimation.

\section{Conclusions and Future Work}
\label{section:conclusion}

This paper presented an end-to-end deep learning based VO and uncertainty estimation method. We utilised a \ac{cnn}-\ac{rnn} hybrid architecture combined with an \ac{mdn}. We evaluated our approach on two public autonomous driving datasets, namely KITTI and nuScenes. Our experiments demonstrate that our method can successfully estimate VO and uncertainty. Furthermore, our model achieves promising results under challenging conditions, such as rain and night scenes, where traditional VO approaches would fail. As future work, we plan to extend our work to utilise multiple cameras and fuse their estimations based on the estimated uncertainties.

{\small
\bibliographystyle{ieee_fullname}
\bibliography{egbib}
}

\end{document}